# Model of Interaction between Learning and Evolution

Vladimir G. Red'ko

Scientific Research Institute for System Analysis, Russian Academy of Sciences, Moscow, Russia
vgredko@gmail.com

**Abstract**

The model of interaction between learning and evolutionary optimization is designed and investigated. The evolving population of modeled organisms is considered. The mechanism of the genetic assimilation of the acquired features during a number of generations of Darwinian evolution is studied. It is shown that the genetic assimilation takes place as follows: phenotypes of modeled organisms move towards the optimum at learning; then the selection takes place; genotypes of selected organisms also move towards the optimum. The hiding effect is also studied; this effect means that strong learning can inhibit the evolutionary search for the optimal genotype. The mechanism of influence of the learning load on the interaction between learning and evolution is analyzed. It is shown that the learning load can lead to a significant acceleration of evolution.

# 1 Introduction

After the appearance of the Darwinian theory of evolution, many researchers asked the following question. The evolutionary processes are based on mutations and further selection. So, are random mutations able to ensure discovering of very non-trivial useful features of living organisms? In the XIX century, the concepts, which suggest that interaction between learning (or other processes of the acquisition of organism features during the life of the organism) and the evolutionary process is possible, appeared (Baldwin, 1896; Morgan, 1896; Osborn, 1896). According to these concepts, learning can contribute significantly to the evolutionary process. This type of influence of learning on the evolutionary process is often called the Baldwin effect (Baldwin, 1896). According to this effect, initially acquired features can become inherited during a number of generations. The evolutionary "reinvention" of useful features, initially obtained by means of learning, is often called the genetic assimilation (Waddington, 1942).

A number of authors analyzed interactions between learning and evolution by means of computer simulations (Belew and Mitchell, 1996; Turney et al., 1996; Hinton and Nowlan, 1987; Mayley, 1997; Ackley and Littman, 1992; Red'ko et al., 2005). In particular, Hinton and Nowlan (1987) demonstrated that learning can guide an evolutionary process to find the optimum. Mayley (1997) investigated different aspects of the interaction between learning and evolution and demonstrated that the hiding effect can take place, if the learning is sufficiently strong. The essence of the hiding effect is as follows: if the learning is enough strong to change the phenotype of the organism and organisms are selected at the evolution in accordance with the phenotype, then the selection can weakly depend on the genotype. The hiding effect significantly reduces the role of the genotype at the evolutionary selection, and the genetic assimilation becomes less pronounced.

In addition, Mayley (1997) investigated the influence of the learning load (the cost of learning) on the interaction between learning and evolution. The learning load means that the process of learning has an additional load for the organism and its fitness is reduced under the influence of this load.

Red'ko et al. (2005) modeled interaction between learning and evolutionary optimization of a neural network control system of autonomous agents. The genetic assimilation of the acquired features of agents was observed during several generations of evolution. In addition, it was demonstrated that learning could significantly accelerate the process of the evolutionary optimization. However, it was difficult to analyze detailed mechanisms of interaction between learning and evolution in that work, because these mechanisms were "hidden" in the dynamics of numerous synaptic weights of agent neural networks.

The current paper uses works (Hinton and Nowlan, 1987; Mayley, 1997) as background. However, that works used rather complex forms of the genetic algorithm (with crossovers), so it was difficult to analyze quantitatively the mechanisms of influence of learning on evolutionary optimization. In contrast to works (Hinton and Nowlan, 1987; Mayley, 1997), the current article uses the quasispecies model proposed by Manfred Eigen (Eigen, 1971; Eigen and Schuster, 1979) and our estimations of the evolutionary rate and the efficiency of evolutionary algorithms (Red'ko and Tsoy, 2005, 2006). The quasispecies model considers the evolution that is based on selection and mutations of organism genotypes (without crossovers) and describes the main properties of the evolutionary process. The use of models and methods of works (Eigen, 1971; Eigen and Schuster, 1979; Red'ko and Tsoy, 2005, 2006) allows getting a better understanding of the mechanisms of interaction between learning and evolution.

The current paper analyzes quantitatively the following main properties of interaction between learning and evolution: 1) the mechanism of the genetic assimilation, 2) the hiding effect, 3) the role of the learning load at investigated processes of learning and evolution.

A short description of the current model was presented in the work (Red'ko, 2013). In contrast to that short work, the current article contains the detailed analysis of interaction between learning and evolution. Additionally, the current paper analyzes the scheme by Hinton and Nowlan (1987) by means of the quasispecies model and characterizes mentioned main properties of interaction between learning and evolution for this scheme.

The paper is organized as follows. Section 2 describes the main model. Section 3 contains the results of computer simulation for this model. Analysis of interaction between learning and evolution within the framework of the scheme by Hinton and Nowlan (1987) by means of the quasispecies model is represented in Section 4. Section 5 concludes the paper. Appendix summarizes briefly the main results of our previous estimations of the efficiency of evolutionary algorithms.

## 2 Description of the Model

The evolving population of modeled organisms is considered. Similar to Hinton and Nowlan (1987), we assume that there is a strong correlation between the genotype and the phenotype of the modeled organisms. We assume that the genotype and the phenotype of the organism have the same form, namely, they are chains; symbols of both chains are equal to 0 or 1. The length of these chains is equal to *N*. For example, we can assume that the genotype encodes a modeled DNA chain, "letters" of which are equal to 0 or 1, and the phenotype determines the neural network of organisms, the synaptic weights of the neural network are equal to 0 or 1 too. The initial synaptic weights (at the birth of the organism) are determined by the genotype (for example, the initial synaptic weights can be equal to the genotype symbols). These weights are adjusted by means of learning during the organism's life.

The evolving population consists of *n* organisms, genotypes of organisms are $\mathbf{S}_{Gk}$, $k = 1,..., n$. The organism genotype $\mathbf{S}_{Gk}$ is a chain of symbols, $S_{Gki}$, $i = 1,..., N$. We assume that the length of chains *N* and the number of organisms in the population *n* are large: $N, n \gg 1$. The values *N* and *n* do not change in the course of evolution. Symbols $S_{Gki}$ are equal to 0 or 1. We assume that *N* is so large that only a small part of possible $2^N$ genotypes can be presented in a particular population: $2^N \gg n$. Typical values *N* and *n* in our computer simulations are as follows: $N \sim n \sim 100$.

The evolutionary process is a sequence of generations. The new generation is obtained from the old one by means of selection and mutations. Genotypes of organisms of the initial generation are random. Organisms inherit the genotypes from their parents, these genotypes do not change during the organism life and are transmitted (with small mutations) to their descendants. Mutations are random changes of symbols $S_{Gki}$.

Phenotypes of organisms $\mathbf{S}_{Pk}$ are chains of symbols $S_{Pki}$, $k = 1,..., n$, $i = 1,..., N$; $S_{Pki} = 0$ or 1. The organism receives the genotype at its birth, the phenotype $\mathbf{S}_{Pk}$ at this time moment is equal to the genotype: $\mathbf{S}_{Pk}(t = 1) = \mathbf{S}_{Gk}$. The lifetime of any organism is equal to *T*. The time is discrete: $t = 1,...,T$. *T* is the duration of the generation. The phenotype $\mathbf{S}_{Pk}$ is modified during the organism life by means of learning.

It is assumed that there is the certain optimal chain $\mathbf{S}_\mathbf{M}$, which is searched for in processes of evolution and learning. Symbols $S_{Mi}$ of this chain are also equal to 0 or 1; the length of the chain $\mathbf{S}_\mathbf{M}$ is *N*. For a particular computer simulation, the chain $\mathbf{S}_\mathbf{M}$ is fixed; symbols of this chain are chosen randomly.

Learning is performed by means of the following method of trial and error. Every time moment *t* each symbol of the phenotype $\mathbf{S}_{Pk}$ of any organism is randomly changed to 0 or 1, and if this new symbol $S_{Pki}$ coincides with the corresponding symbol $S_{Mi}$ of the optimal chain $\mathbf{S}_\mathbf{M}$, then this symbol is fixed in the phenotype $\mathbf{S}_{Pk}$, otherwise, the old symbol of the phenotype $\mathbf{S}_{Pk}$ is restored. The probability of the random changing of a symbol during learning is equal to $p_l$. So, during learning, the phenotype $\mathbf{S}_{Pk}$ moves towards the optimal chain $\mathbf{S}_\mathbf{M}$.

At the end of the generation, the selection of organisms in accordance with their fitness takes place. The fitness of *k*-th organism is determined by the final phenotype $\mathbf{S}_{Pk}$ at the time moment $t = T$. We denote this chain $\mathbf{S}_{Fk}$, i.e. we set $\mathbf{S}_{Fk} = \mathbf{S}_{Pk}(t = T)$. The fitness of *k*-th organism is determined by the Hamming distance $\rho = \rho(\mathbf{S}_{Fk}, \mathbf{S}_\mathbf{M})$ between the chains $\mathbf{S}_{Fk}$ and $\mathbf{S}_\mathbf{M}$:

$$f_k = \exp[-\beta \rho(\mathbf{S}_{Fk}, \mathbf{S}_\mathbf{M})] + \varepsilon, \quad (1)$$

where $\beta$ is the positive parameter, which characterizes the intensity of selection, $0 < \varepsilon \ll 1$. The role of the value $\varepsilon$ in (1) can be considered as the influence of random factors of the environment on the fitness of organisms.

The selection of organisms into a new generation is made by means of the well-known method of fitness proportionate selection (or roulette wheel selection). In this method, organisms are selected into a new generation probabilistically. The choice of an organism into the next generation takes place *n* times, so the number of organisms in the population at all generations is equal to *n*. The probability of the selection of *k*-th organism into the next generation at a particular choice is equal to

$$q_k = \frac{f_k}{\sum_{j=1}^{n} f_j}.$$

Therefore, at any choice, the probability of the selection of a particular organism into the next generation is proportional to its fitness.

Thus, organisms are selected at the end of a generation in accordance with their final phenotypes $S_{Fk} = S_{Pk}(t = T)$, i.e. in accordance with the final result of learning, whereas genotypes $S_{Gk}$ (modified by small mutations) are transmitted from parents to descendants.

As descendants of organisms obtain genotypes $S_{Gk}$ that organisms received from their parents and not phenotypes $S_{Pk}$, the evolutionary process has Darwinian character.

Additionally, similar to Mayley (1997), we take into account the learning load (the cost of learning), namely, we assume that the learning process has a certain burden on the organism and the fitness of the organism may be reduced under the influence of the load. For this purpose, we consider the modified fitness of organisms:

$$f_{mk} = \exp(-\alpha d) \{\exp[-\beta\rho(S_{Fk}, S_M)] + \varepsilon\}, \qquad (2)$$

where $\alpha$ is the positive parameter, which takes into account the learning load, $d = \rho(S_{Gk}, S_{Fk})$ is the Hamming distance between the initial $S_{Pk}(t = 1) = S_{Gk}$ and the final phenotype $S_{Pk}(t = T) = S_{Fk}$ of the organism, i.e. the value that characterizes the intensity of the whole learning process of the organism during its life.

It should be noted that since genotypes $S_{Gk}$ of the organisms in the initial population are random, the average Hamming distance between these chains and the optimal one $S_M$ is equal to $N/2$. The chains $S_k$ should overcome this distance at learning and evolution in order to reach $S_M$.

## 3 Results of Computer Simulation

### 3.1 Scheme and Parameters of Simulation

Two modes of operation of the model are consider below: 1) the regime of the evolution combined with learning, as described above, 2) the regime of "pure evolution", that is the evolution without learning, in this case, the learning does not occur and $S_{Pk} = S_{Gk}$. Additionally, the influence of the learning load is analyzed, in this case, the fitness of an organism is calculated according to (2). The model is investigated by means of computer simulation.

The parameters of the model at simulation are chosen in such manner that the evolutionary search is effective; the experience of the work (Red'ko and Tsoy, 2005) for the case of pure evolution is used at this choice. The fitness of the organisms in that work was determined analogously to the expression (1), only the influence of random factors was not taken into account (formally this means that the value $\varepsilon$ was equal to 0).

The choice of parameters for the current simulation is as follows. We believe that the length of the chains is sufficiently large: $N = 100$. We also set $\beta = 1$, this corresponds to a sufficiently high intensity of selection, so the selection time is small, thus the time of the evolutionary search is determined mainly by the intensity of mutations. On the one hand, the intensity of mutations must not be too large, in order to remove the possibility of mutational losses of already found good organisms. On the other hand, the intensity of mutations must not be too small, in order to ensure the sufficiently intensive mutational search during the evolutionary optimization. Taking this into account, we believe that the probability to change any symbol in any chain $S_{Gk}$ at one generation at mutations is $p_m = N^{-1} = 0.01$. At this mutation intensity $p_m$ approximately one symbol in the genotype of any organism is changed at one generation, i.e. during one generation, the Hamming distance $\rho$ between genotypes $S_{Gk}$ of organisms and the optimal chain $S_M$ changes on average by 1 by means of mutations. The selection leads to a decrease of the distance $\rho$. Since the intensity of the selection is large, and the Hamming distance between genotypes $S_{Gk}$ in the initial population and the optimal chain $S_M$ is of the order of $N$, the whole process of the evolutionary optimization takes approximately $G_T \sim N$ generations. This estimation of the evolutionary rate is true, if the population size is sufficiently large and the fluctuation effects and the neutral selection of organisms (that is the selection independent on the fitness of organisms) can be neglected. To satisfy this condition, it is enough to require that the characteristic time (a number of generations) of the neutral selection, which is of the order of the population size $n$ (Kimura, 1983; Red'ko and Tsoy, 2005), should be greater or of the order of $G_T$, so we believe that $n = N$ (so, $n \sim G_T \sim N$). See also Appendix for more details.

Thus, the parameters of simulation in accordance with the experience of the work (Red'ko and Tsoy, 2005) are chosen as follows: $N = 100$, $\beta = 1$, $p_m = N^{-1} = 0.01$, $n = N = 100$.

In the current model we also believe that the probability of a random replacement of any symbol during learning $p_l$ is rather large: $p_l \sim 1$, the number of time moments during any generation $T$ is equal to 2 (choice of such parameters $p_l$ and $T$ means that learning is rather strong), the parameter $\varepsilon$ is small: $\varepsilon = 10^{-6}$. The majority of simulations are carried out at $p_l = 1$, only in one case the value $p_l$ is equal to 0.5.

The results of simulation are averaged over 1000 or 10000 calculations corresponding to different random seeds. This averaging insures good accuracy of simulation; typical errors are smaller than 1-2%. The results of simulation are described below.

## 3.2 Comparison of Regimes of Pure Evolution and Evolution Combined with Learning

Figure 1 shows the dependence of the average Hamming distance $\rho = \rho(S_{Gk},S_M)$ between genotypes $S_{Gk}$ of organisms in the population and the optimal chain $S_M$ on the generation number $G$. The curve 1 characterizes the regime of evolution combined with learning; the curve 2 characterizes the regime of pure evolution. The dependences are averaged over 1000 calculations. The fitness of organisms is determined by the expression (1). We can see that the pure evolution without learning (the curve 2) does not optimize organisms $S_k$ at all; whereas evolution combined with learning (the curve 1) obviously ensures the movement towards the optimal chain $S_M$. Errors of values $<\rho>$ at the plots are smaller than 0.3.

To understand, why the pure evolution does not ensure a decrease of the value $\rho$, let us estimate the value of the fitness (1) in the initial population. The Hamming distance $\rho = \rho(S_{Gk},S_M)$ for initial genotypes is of the order of $N/2 = 50$, therefore, $\exp(-\rho) \sim 10^{-22}$ and $\exp(-\rho) \ll \varepsilon$. This means that all organisms of the population have approximately the same value of the fitness $f_k \approx \varepsilon$. Consequently, the evolutionary optimization of genotypes does not occur in the case of the pure evolution. Thus, the movement towards $S_M$ occurs only in the presence of learning; this movement leads to the decrease of the value $\rho$. A similar influence of learning on the evolutionary optimization (though in another context) was described by Hinton and Nowlan (1987).

Let us consider the effect of the acceleration of the evolutionary process by learning (the curve 1 in Figure 1). Analysis of the results of simulations shows that the gradual decrease of the values $\rho = \rho(S_{Gk},S_M)$ occurs as follows. First, the learning shifts the distribution of organisms $n(\rho)$ on the value $\rho$ towards smaller $\rho$, so the values $\rho = \rho(\mathbf{S}_{Fk},\mathbf{S}_M)$ become small enough, such that $\exp[-\rho(\mathbf{S}_{Fk},\mathbf{S}_M)]$ is of the order of $\varepsilon$. Consequently, the fitnesses of organisms in the population in accordance with (1) become essentially different; so organisms with small values $\rho(\mathbf{S}_{Fk},\mathbf{S}_M)$ are selected into the population of the next generation. It is intuitively clear that the genotypes of $\mathbf{S}_{Gk}$ of selected organisms should be rather close to the final phenotypes $\mathbf{S}_{Fk}$ (obtained as a result of the learning) of these organisms. Thus, the result of the selection is choosing of organisms, which genotypes are also moving to the optimal chain $\mathbf{S}_M$. Therefore, values $\rho$ in the new population decrease.

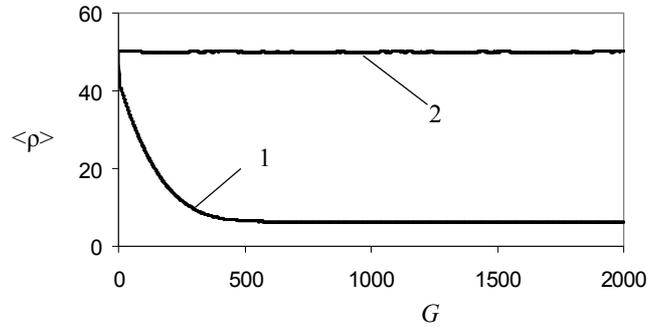

Figure 1: The dependence of the average Hamming distance $\langle\rho\rangle = \langle\rho(\mathbf{S}_{Gk},\mathbf{S}_M)\rangle$ between genotypes $\mathbf{S}_{Gk}$ and the optimal chain $\mathbf{S}_M$ on the generation number $G$. The curve 1 characterizes the regime of evolution combined with learning; the curve 2 characterizes the regime of pure evolution. Results are averaged over 1000 calculations.

The described mechanism of the genetic assimilation is characterized by Figure 2, which shows the distributions of the number of organisms $n(\rho)$ for given $\rho$ in the population for different moments of the first generation. The curve 1 shows the distribution $n(\rho)$ for $\rho = \rho(\mathbf{S}_{Gk},\mathbf{S}_M)$ for the initial genotypes of organisms at the beginning of the generation. The curve 2 shows the distribution $\rho = \rho(\mathbf{S}_{Fk},\mathbf{S}_M)$ for organisms after the learning, but before the selection. The curve 3 shows the distribution $\rho = \rho(\mathbf{S}_{Fk},\mathbf{S}_M)$ for organisms, selected in accordance with the fitness (1). The curve 4 shows the distribution $\rho = \rho(\mathbf{S}_{Gk},\mathbf{S}_M)$ for the genotypes of selected organisms at the end of the generation. The genotypes of selected organisms $\mathbf{S}_{Gk}$ are sufficiently close to the final phenotypes of learned and selected organisms $\mathbf{S}_{Fk}$, therefore the distribution $\rho = \rho(\mathbf{S}_{Gk},\mathbf{S}_M)$ for genotypes (the curve 4) moves towards the distribution for final phenotypes $\mathbf{S}_{Fk}$ (the curve 3). Similar displacement of the distribution $n(\rho)$ towards smaller values $\rho$ takes place in the next generations. Errors of values $n(\rho)$ at the plots are smaller than 0.3.

Such displacement reveals the mechanism of reduction of $\langle\rho\rangle$ in the presence of learning: the selection leads to the genotypes of organisms $\mathbf{S}_{Gk}$, which are closer to the phenotypes of learned and selected organisms $\mathbf{S}_{Fk}$, than the initial genotypes of organisms at the beginning of the generation. Consequently, the transition from the curve 1 to the curve 4, i.e. the decrease of the values $\rho$, takes place during the generation.

It should be underlined that the decrease of values $\rho$ at learning should be sufficiently large in order to ensure the small role of the parameter $\varepsilon$ and the significant difference of the fitnesses (1) of organisms after the learning, and therefore, the effective selection of organisms with small values $\rho(\mathbf{S}_{Fk},\mathbf{S}_M)$. This selection corresponds to the essential decrease of values $\rho$ at the transition from the curve 2 to the curve 3 in Figure 2. It is clear that in order to guarantee the effective operation this mechanism, the learning should be enough strong. The other role of strong learning is characterized in the next subsection.

It should be noted that the displacement of the distribution $n(\rho)$ at learning in the first generation can be estimated as follows. Before learning, the value $\rho(S_{Pk},S_M)$ (the number of symbols of phenotype $S_{Pk}$ that do not coincide with corresponding symbols of the optimal chain $S_M$) is approximately equal to $N/2 = 50$. After the first step of learning approximately a half of non-coinciding symbols are changed ($p_l = 1$), so the value $\rho(S_{Pk},S_M)$ becomes to be approximately equal to $N/4 = 25$. After the second step of learning (at the end of the generation) the next half of non-coinciding symbols are changed, so the value $\rho(S_{Pk},S_M)$ diminishes to $N/8 = 12.5$. This is in agreement with the curve 2 in Figure 2.

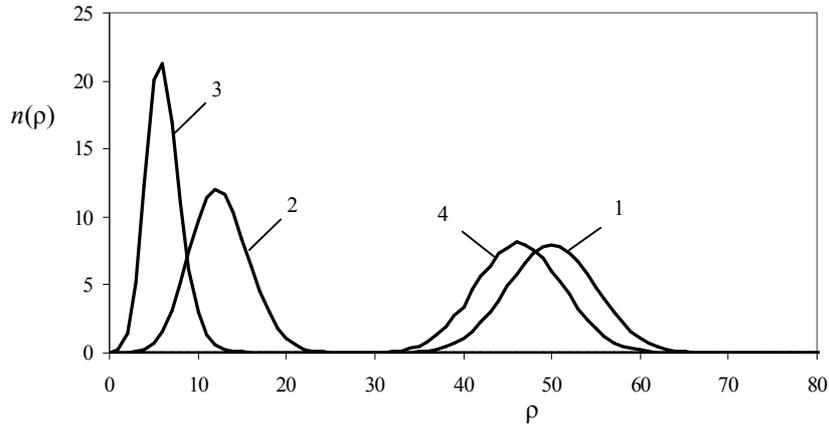

Figure 2: The distributions $n(\rho)$ in the first generation of evolution for different moments of the generation. The curve 1 is the distribution $n(\rho)$ for $\rho = \rho(S_{Gk},S_M)$ for the original genotypes before learning. The curve 2 is the distribution $n(\rho)$ for $\rho = \rho(S_{Fk},S_M)$ for organisms after the learning, but before the selection. The curve 3 is the distribution $n(\rho)$ for $\rho = \rho(S_{Fk},S_M)$ for selected organisms. The curve 4 is the distribution $n(\rho)$ for $\rho = \rho(S_{Gk},S_M)$ for the genotypes of selected organisms at the end of the generation. Results are averaged over 10000 calculations.

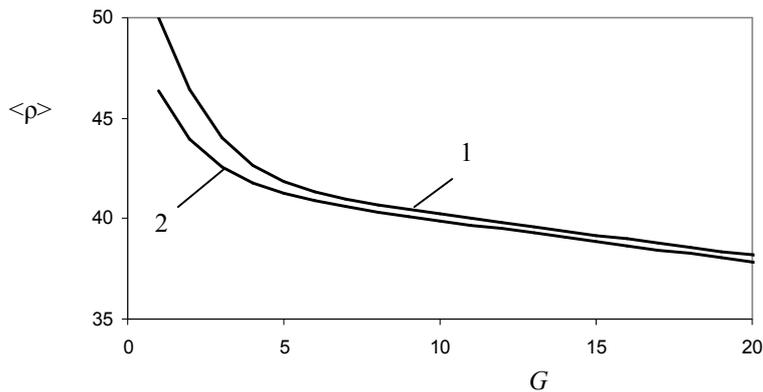

Figure 3: The dependence of $<\rho> = <\rho(S_{Gk},S_M)>$ for the genotypes of organisms on the generation number $G$ for different moments of generations: for the beginning of the generations (1) and for the end of the generations (2). Results are averaged over 10000 calculations.

The same mechanism of decreasing the value ρ for the regime of evolution combined with learning is illustrated by Figure 3. This figure shows the dependence of the average distance <ρ> = <ρ($S_{Gk}$,$S_M$)> between the genotypes of organisms of the population $S_{Gk}$ and the optimal chain $S_M$ on the generation number $G$ for the moments of the beginning of the generations (the curve 1) and for the moments after selection (the curve 2). Figure 3 demonstrates that at the end of the generation (after the selection) the average value ρ is clearly decreased as compared with the beginning of the generation. The value of this decrease of <ρ> is maximal at the first generations, whereas the amount of the decrease becomes smaller at the next generations.

The described results show that learning can lead to the effective genetic assimilation and to the radical acceleration of the evolutionary search.

### 3.3 Hiding Effect

Thus, the strong learning can accelerate the evolutionary search. However, the strong learning can also prevent a finding of the optimal genotype. The curve 1 in Figure 1 shows that at large $G$ the decrease of values <ρ> = <ρ($S_{Gk}$,$S_M$)> is limited: the final value <ρ> remains quite large, the asymptotic value <ρ> is approximately equal to 6.2. This is due to the fact that at large $G$ ($G$ ~1000) the strong learning ($p_l = 1$, $T = 2$) results in finding the optimal phenotype $S_{Popt} = S_M$ independently on the genotype $S_{Gk}$. Therefore, at the final stages of the evolutionary process, the genotypes $S_{Gk}$ do not move towards the optimum $S_M$. So, the hiding effect (Mayley, 1997) is observed.

Figure 4 characterizes the mechanism of the hiding effect. This figure represents the distributions $n(ρ)$ at the end of the evolutionary process (at $G = 2000$) for different moments of the generation. The results are for the described case of simulation for the regime of evolution combined with learning. Figure 4 shows that the distribution $n(ρ)$ after the learning includes organisms, for which ρ($S_{Fk}$,$S_M$) = 0, i.e. the optimal phenotype $S_{Popt} = S_M$ is found by means of the learning. Though the selection in accordance with values ρ($S_{Fk}$,$S_M$) occurs, the distance between the initial genotype distribution (the curve 1) and the final genotype distribution (the curve 4) is sufficiently small. Therefore, further reduction of ρ = ρ($S_{Gk}$,$S_M$) at the end of the evolutionary process does not occur. The hiding effect is confirmed by the fact that at the end of the evolution the curves (that are shown in Figure 4) do not shift for successive generations. This effect is also consistent with the fact that the value <ρ> = <ρ($S_{Gk}$,$S_M$)> becomes constant at large $G$ (see the curve 1 in Figure 1). The distributions $n(ρ)$ for genotypes at the beginning of the generation and after the selection (curves 1 and 4 in Figure 4) differ slightly, this is due to mutations that lead to a small increase of ρ in the beginning of a generation as compared with the distribution after selection. Thus, at the end of the evolutionary process, the strong learning results in finding of the optimal phenotype; hence a further optimization of genotypes does not occur.

The hiding effect can be substantially relaxed by reducing the intensity of learning. The dependence of <ρ> = <ρ($S_{Gk}$,$S_M$)> for genotypes on the generation number $G$ for the weakened learning ($p_l = 0.5$) is represented in Figure 5. For this case, the rate of the decrease of the value <ρ> during the evolutionary process is smaller as compared with the previous result (Figure 1, the curve 1); however, the final value <ρ> = <ρ($S_{Gk}$,$S_M$)> is essentially reduced and becomes approximately equal to 1.4. Consequently, the weakening of the learning leads to the fact that the phenotype, which determines the selection, in the greater degree depends on the genotype $S_{Gk}$; so, the selection of organisms having genotypes, which are quite close to $S_M$, takes place.

The hiding effect can be eliminated in another way: the learning process can be turned off at large $G$. Figure 6 shows the simulation result, for which the learning is turned off at $G =$ 1000. Simulation parameters are the same as for the calculation represented in Figure 1 (the curve 1). Turning off the learning results in the sudden decrease of the value $<\rho>$ = $<\rho(S_{Gk},S_M)>$ immediately after the generation $G = 1000$, this has the following explanation. As at $G = 1000$, the value $<\rho(S_{Gk},S_M)>$ is approximately equal to 6, then for this population we have $\exp(-\rho) \sim 0.001 \gg \varepsilon$, consequently, fitnesses of the organisms (calculated according to (1)) are essentially different. Therefore, the evolutionary optimization of genotypes is successfully functioning; then the evolutionary process leads to the effective finding of the optimal genotype $S_{Gopt} = S_M$.

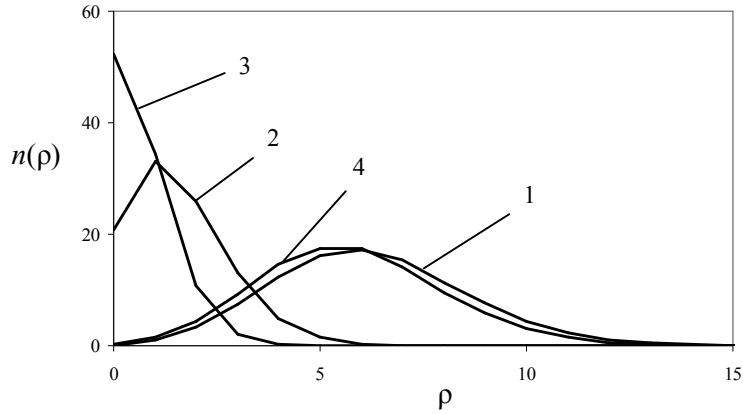

Figure 4: The distributions $n(\rho)$ at the end of the evolutionary process (at $G = 2000$) for different moments of the generation. The curve 1 is the distribution of $\rho = \rho(S_{Gk},S_M)$ for the initial genotypes before learning. The curve 2 is the distribution of $\rho = \rho(S_{Fk},S_M)$ for organisms after the learning, but before the selection. The curve 3 is the distribution of $\rho = \rho(S_{Fk},S_M)$ for selected organisms. The curve 4 is the distribution of $\rho = \rho(S_{Gk},S_M)$ for the genotypes of selected organisms at the end of the generation. Results are averaged over 1000 calculations.

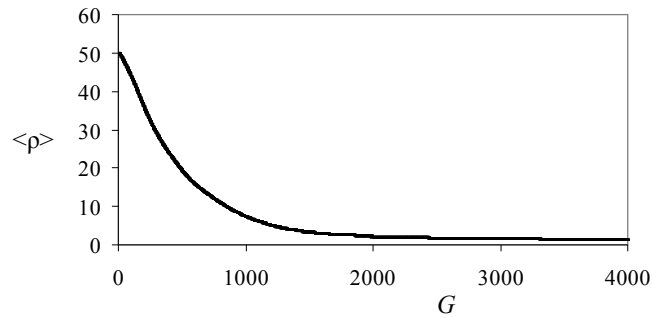

Figure 5: The dependence of $<\rho>$ = $<\rho(S_{Gk},S_M)>$ on the generation number $G$ for the case of the weakened learning: $p_l = 0.5$ (results are averaged over 1000 calculations); as compared with the case of $p_l = 1$, the evolutionary rate is reduced, but genotypes of organisms, which are essentially closer to $S_M$, are found.

Thus, the mechanism of the hiding effect is analyzed. This effect means that the strong leaning prevents a finding of the optimal genotype, as such learning increases the chances of finding a good phenotype independently on the genotype of the organism. In our case, the hiding effect is observed at the end of the evolutionary process.

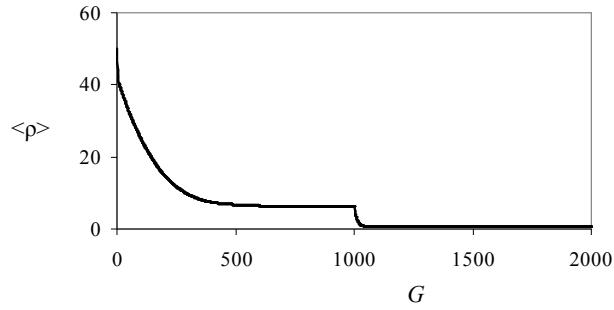

Figure 6: The dependence of $<\rho> = <\rho(\mathbf{S}_{Gk},\mathbf{S}_{M})>$ on the generation number $G$; the learning is turned off at $G = 1000$, then the evolutionary process leads to the effective finding of the optimal genotype (results are averaged over 1000 calculations).

### 3.4 Influence of the Learning Load on the Modeled Processes

We also analyzed the influence of the learning load on the modeled processes. For this case, the fitness of organisms is determined by the expression (2). The simulation is performed for the mentioned parameters ($N = n = 100$, $\beta = 1$, $p_m = 0.01$, $p_l = 1$, $T = 2$, $\varepsilon = 10^{-6}$), the value $\alpha$ is equal to 1. The simulation results are represented in Figures 7, 8. Figure 7 shows the dependence of the average Hamming distance $<\rho> = <\rho(\mathbf{S}_{Gk},\mathbf{S}_{M})>$ between genotypes $\mathbf{S}_{Gk}$ and the optimal chain $\mathbf{S}_{M}$ on the generation number $G$. Figure 8 shows the distributions $n(\rho)$ of values $\rho$ for different moments of the first generation of the evolution.

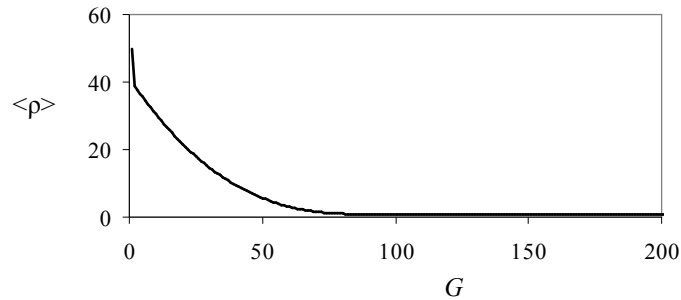

Figure 7: The dependence of $<\rho> = <\rho(\mathbf{S}_{Gk},\mathbf{S}_{M})>$ on generation number $G$; the influence of the learning load is considered; the fitness of organisms is determined by the expression (2); the decrease of values $<\rho>$ is much faster than that of in Figure 1 (results are averaged over 1000 calculations).

The comparison of Figures 1, 2 and Figures 7, 8 shows that the learning load leads to the considerable acceleration of the evolutionary search for the optimal chain $\mathbf{S}_{M}$. This acceleration is due to the fact that the learning load results in the more strong selection of organisms that have small distance $\rho(\mathbf{S}_{Gk},\mathbf{S}_{Fk})$ between the initial $\mathbf{S}_{Pk}(t = 1) = \mathbf{S}_{Gk}$ and the final $\mathbf{S}_{Pk}(t = T) = \mathbf{S}_{Fk}$ phenotypes, than for the case of the fitness (1). This form of the selection in accordance with the expression (2) leads to the additional minimization of changes of phenotypes $S_{Pk}$ during the learning process. The distribution 3 in Figure 8 has some "extended tail" to the right; this is in accordance with the minimization of changes of phenotypes $S_{Pk}$ during the learning.

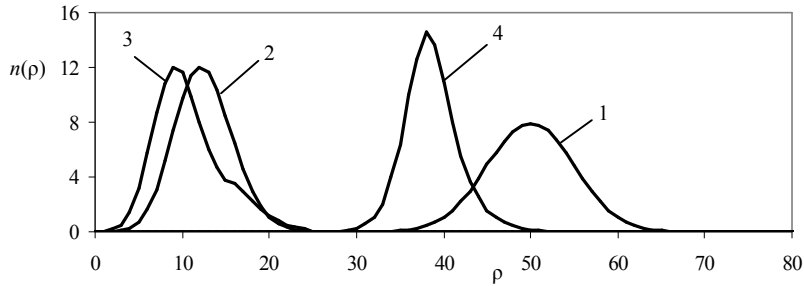

Figure 8: The distributions $n(\rho)$ for different moments of the first generation of evolution; the learning load is taken into account; the fitness of organisms is determined by the expression (2). The curve 1 is the distribution of $\rho = \rho(S_{Gk}, S_M)$ for the original genotypes before learning. The curve 2 is the distribution of $\rho = \rho(S_{Fk}, S_M)$ for organisms after the learning, but before the selection. The curve 3 is the distribution of $\rho = \rho(S_{Fk}, S_M)$ for selected organisms. The curve 4 is the distribution of $\rho = \rho(S_{Gk}, S_M)$ for the genotypes of selected organisms at the end of the generation. The displacement of the distribution 4 to smaller values $\rho$ is significantly larger than in Figure 2. Results are averaged over 10000 calculations.

Figure 9 represents the distributions $n(\rho)$ at the end of the evolutionary process (at $G = 200$) for different moments of the generation. This figure shows that the optimal genotype $S_{Gopt} = S_M$ in the considered case is found.

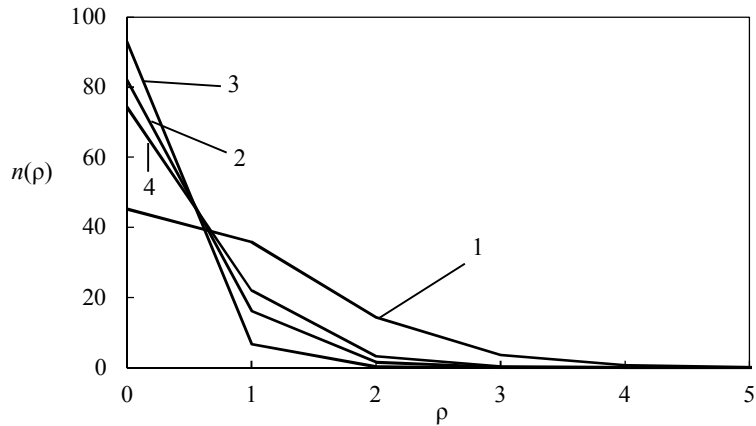

Figure 9: The distributions $n(\rho)$ at the end of evolution (at $G = 200$) for different moments of the generation; the learning load is taken into account; the fitness of organisms is determined by the expression (2). The curve 1 is the distribution of $\rho = \rho(S_{Gk}, S_M)$ for the initial genotypes before learning. The curve 2 is the distribution of $\rho = \rho(S_{Fk}, S_M)$ for organisms after the learning, but before the selection. The curve 3 is the distribution of $\rho = \rho(S_{Fk}, S_M)$ for selected organisms. The curve 4 is the distribution of $\rho = \rho(S_{Gk}, S_M)$ for the genotypes of selected organisms at the end of the generation. Results are averaged over 1000 calculations.

It should be underlined that the genetic assimilation for cases of the fitness, which is determined by the expression (1) and the expression (2), has the same nature. In both cases, genotypes of selected organisms $S_{Gk}$ approach to final phenotypes $S_{Fk}$ of learned and selected organisms. That is in both Figure 2 and Figure 8 the curve 4 moves towards the curve 3. A significant difference consists only in the fact that the learning load makes this movement more evident and more effective. Thus, the learning load leads to more effective optimization of genotypes of $S_{Gk}$; and consequently, the evolution process is significantly accelerated. Figure 9 demonstrates that the learning load results in finding of the optimal genotype $S_{Gopt} = S_M$. The learning load makes the genetic assimilation more profound. The hiding effect is absent in this case.

Thus, the computer simulation shows that the genetic assimilation, the hiding effect, and the significant acceleration of the genetic assimilation and the evolutionary process under the influence of the learning load are observed in the current model.

Some issues and possible variations of the main model are described in subsection 3.5 and 3.6.

### 3.5 Probabilistic and Deterministic Selection

The considered model uses the probabilistic selection of individuals in accordance with their fitness; the method of fitness proportionate selection is used. Therefore, the presence of the small parameter $\varepsilon$ in expressions (1) and (2) leads to the fact that a purely evolutionary process did not ensure finding the optimal sequence $S_M$. It is possible to use the deterministic selection instead of the probabilistic one. For example, we can calculate the fitness of all organisms in a computer program and select into the next generation exactly the half of the individuals, which have larger fitness as compared with the rest of organisms of the population, and duplicate selected organisms. We have executed the simulation for this case of the deterministic selection. The simulation showed that in this case, the pure evolution leads to finding the optimal genotype $S_{Gopt} = S_M$; the characteristic time of convergence of the evolutionary process is of the order of $N$ generations. However, the deterministic selection implies that the fitness of the individuals (1) is calculated with great accuracy in a computer program; this is unnatural for real biological processes. For the biological processes, it is more natural to suppose that the selection has the probabilistic character, as it is assumed above.

### 3.6 Variant of the Model

Expressions (1) and (2) for fitness assume that possible random variations of environment are considered by means of the parameter $\varepsilon$. This is the simplest form of such consideration. We can consider the influence of possible random variations of environment more directly. In this subsection, we define the fitness of organisms, slightly modifying expressions (1) and (2):

$$f_k = \exp[-\beta\rho(S_{Fk},S_M)] + 2\varepsilon\xi , \qquad (1a)$$

$$f_{mk} = \exp(-\alpha d) \{\exp[-\beta\rho(S_{Fk},S_M)] + 2\varepsilon\xi\} , \qquad (2a)$$

where $\xi$ is the random variable uniformly distributed in the interval [0,1]. The values $\xi$ are different for different moments of fitness calculations. The expression (2a) takes into account the learning load.

Using the expressions (1a) and (2a), we reproduced all described computer simulations for this variant of the model (results of simulations were averaged over large number of calculations). The results of these simulations actually coincided with the results described above; the difference between values <ρ> and $n(\rho)$ for two types of simulations are smaller than 0.3, that is this difference is no greater than errors of simulations. This coincidence is due to the averaging procedure for these two variants of the model.

## 4 Comparison with the Approach by Hinton and Nowlan

This section uses the approach by Hinton and Nowlan (1987) as well as the quasispecies model (Eigen, 1971; Eigen and Schuster, 1979). We consider the additional model that is very similar to the main model described above. The additional model is based on the approach by Hinton and Nowlan (1987). Almost all assumptions of the additional model are the same as in the main model. In the additional model, we suppose that organisms of the evolving population have genotypes $\mathbf{S}_{Gk}$ and phenotypes $\mathbf{S}_{Pk}$, $k = 1,..., n$. $\mathbf{S}_{Gk}$ and $\mathbf{S}_{Pk}$ are chains of symbols, $S_{Gki}$, $S_{Pki}$, $i = 1,..., N$, $N$, $n \gg 1$. Symbols $S_{Gki}$, $S_{Pki}$ are equal to 0 or 1. $\mathbf{S}_{Pk}(t=1) = \mathbf{S}_{Gk}$, $t = 1,...,T$. $T$ is the duration of the generation. There is the certain optimal chain $\mathbf{S_M}$ (components of which $S_{Mi}$ are equal to 0 or 1, $i = 1,..., N$), which is searched for in the process of evolution and learning. Learning is performed by means of the method of trial and error (as described above). At the end of the generation, the selection of organisms in accordance with their fitness takes place; the method of fitness proportionate selection is used.

Only the fitness of organisms in the additional model is defined in another way, as follows.

1) If learning takes place, the fitness of $k$-th organism is determined by the final phenotype $\mathbf{S}_{Pk}$ at $t = T$:

$$f_k = \exp[-\beta\rho(\mathbf{S}_{Fk},\mathbf{S_M})] \,, \tag{3a}$$

where $\mathbf{S}_{Fk} = \mathbf{S}_{Pk}(t=T)$, $\rho = \rho(\mathbf{S}_{Fk},\mathbf{S_M})$ is the Hamming distance between $\mathbf{S}_{Fk}$ and $\mathbf{S_M}$.

2) If there is no learning, then the fitness is:

$$f_k = \begin{cases} 1, \text{if } \mathbf{S}_{Gk} = \mathbf{S_M} \\ 0, \text{if } \mathbf{S}_{Gk} \neq \mathbf{S_M} \end{cases} . \tag{3b}$$

The additional model has been analyzed by means of computer simulation. All simulations have been made for the case, when the learning takes place; that is the fitness is determined mainly by the expression (3a). Additionally, the influence of the leaning load is taken into account. In this case, the fitness is modified:

$$f_{mk} = \exp(-\alpha d) \exp[-\beta\rho(\mathbf{S}_{Fk},\mathbf{S_M})] \,, \tag{4}$$

where $d = \rho(\mathbf{S}_{Gk},\mathbf{S}_{Fk})$.

The results for the additional model are almost the same as the described results for the main model. The genetic assimilation, the hiding effect, and the influence of the leaning load are observed in the case of the additional model.

For example, Figure 10 shows the dependence of the average Hamming distance <ρ> = <ρ($\mathbf{S}_{Gk},\mathbf{S_M}$)> between genotypes $\mathbf{S}_{Gk}$ of organisms in the population and the optimal chain $\mathbf{S_M}$ on the generation number $G$. The parameters of simulation are: $N = n = 100$, $\beta = 1$, $p_m = 0.01$, $p_l = 1$, $T = 2$.

Figure 10 shows that the dependence of <ρ> on $G$ is almost the same as the curve 1 in Figure 1. The distributions $n(\rho)$ in the population for different moments of the first generation are very close to those ones shown in Figure 2. These results demonstrate that the genetic assimilation is definitely observed in the additional model.

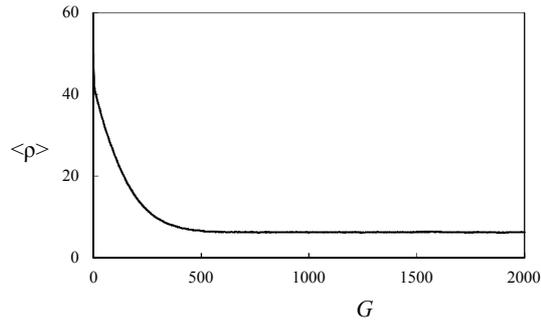

Figure 10: The dependence of the average Hamming distance $\langle\rho\rangle = \langle\rho(\mathbf{S}_{Gk},\mathbf{S}_M)\rangle$ between genotypes $\mathbf{S}_{Gk}$ of organisms and the optimal chain $\mathbf{S}_M$ on the generation number $G$; the fitness of organisms is determined by the expression (3a). Results are averaged over 1000 calculations.

According to Figure 10, the asymptotic value $\langle\rho\rangle$ at large $G$ is approximately equal to 6.2. The distributions $n(\rho)$ in the population for different moments of the generation at the end of the evolutionary process are almost identical to the distributions in Figure 4. So, the hiding effect is also observed in the additional model.

Only in the case of simulations corresponding to the influence of the loading load, there is a small difference for two considered models. In this case, the simulation for the additional model is performed for the parameters $N = n = 100$, $\beta = 1$, $p_m = 0.01$, $p_l = 1$, $T = 2$, $\alpha = 1$. The fitness is determined by the expression (4). The dependence of the average distance $\langle\rho\rangle = \langle\rho(\mathbf{S}_{Gk},\mathbf{S}_M)\rangle$ between genotypes $\mathbf{S}_{Gk}$ and the optimal chain $\mathbf{S}_M$ on the generation number $G$ is very close to that of shown in Figure 7. Figure 11 shows the distributions $n(\rho)$ for different moments of the first generation of the evolution. This figure demonstrates that the displacement of the distributions $n(\rho)$ is similar to that of shown on Figure 8, however, there is a small difference between these displacements. Nevertheless, the role of the loading load in the additional model is the same as in the main model. In particular, the loading load leads to the effective genetic assimilation and the significant acceleration of the evolutionary optimization.

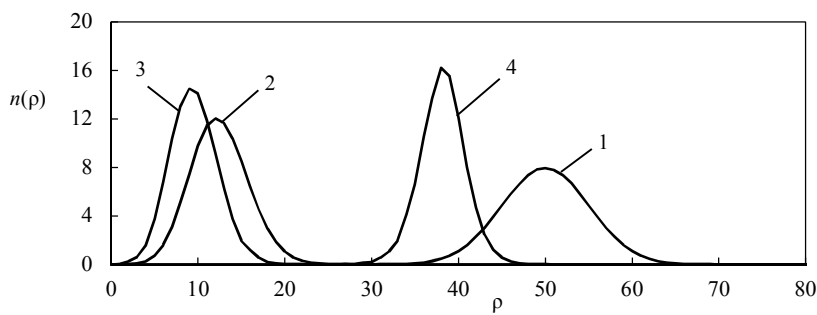

Figure 11: The distributions $n(\rho)$ in the first generation of evolution; the learning load is taken into account; the fitness of organisms is determined by the expression (4). The curve 1 is the distribution of $\rho = \rho(\mathbf{S}_{Gk},\mathbf{S}_M)$ for the original genotypes before learning. The curve 2 is the distribution of $\rho = \rho(\mathbf{S}_{Fk},\mathbf{S}_M)$ for organisms after the learning, but before the selection. The curve 3 is the distribution of $\rho = \rho(\mathbf{S}_{Fk},\mathbf{S}_M)$ for selected organisms. The curve 4 is the distribution of $\rho = \rho(\mathbf{S}_{Gk},\mathbf{S}_M)$ for the genotypes of selected organisms at the end of the generation. Results are averaged over 10000 calculations.

Analogously to Hinton and Nowlan (1987), we can estimate the efficiency of influence of learning on the evolutionary optimization as follows. The dependence of the value $<\rho> = <\rho(\mathbf{S}_{Gk},\mathbf{S_M})>$ on the generation number $G$ in the considered case is very close to the plot shown in Figure 7. The number of generations needed to find the optimal genotype is of the order of 100. The total number of organisms participating in the evolutionary search is of the order of $10^4$. This value is radically smaller than the number of organisms needed for finding the optimum without leaning, at random search (see the expression (3b)), which can be estimated by the value $2^{100} \sim 10^{30}$.

The coincidence of the essential results for the main and additional models shows that the role of the parameter $\varepsilon$ (see expressions (1) and (2)) in the main model is rather small. This parameter is essentially significant only for clear comparison of regimes of pure evolution and evolution combined with learning (see Figure 1).

Thus, the comparison with the approach by Hinton and Nowlan (1987) demonstrates that in the framework of this approach, we can design the model, which reveals actually the same properties of interaction between learning and evolution as the main model.

The analysis of both models shows that a) the genetic assimilation, b) the hiding effect, and c) the significant acceleration of the genetic assimilation and the evolutionary process under the influence of the leaning load are observed in these models under the following assumptions:
1) Each organism of the evolving population has a genotype and a phenotype.
2) The genotype and the phenotype are chains of symbols; the both chains have the same form.
3) Genotypes of organisms are transmitted from parents to descendants with small mutations. The genotype of the organism is not changed during its life.
4) The initial phenotype of the organism at its birth is equal to the organism genotype.
5) There is a certain optimal chain, which is searched for by means of learning and evolution. The optimal chain has the same form as the genotype and the phenotype.
6) The phenotype is essentially adjusted by means of learning during the organism lifetime. During learning, the phenotype moves towards the optimal chain.
7) The selection of organisms into a new generation occurs in accordance with final phenotypes of organisms.

# 5 Conclusion

Thus, the model of interaction between learning and evolution has been constructed and investigated.

The mechanism of the genetic assimilation is studied in detail. It is shown that the genetic assimilation takes place as follows. The phenotypes of modeled organisms move towards the optimum at learning; then the selection in accordance with final phenotypes takes place; the genotypes of selected organisms also move towards the optimum. It is shown that the genetic assimilation can lead to a radical acceleration of the evolutionary search.

The mechanism of the hiding effect is analyzed. This effect means that strong learning inhibits the evolutionary search for the optimal genotype, if this learning increases the chances of finding a good phenotype regardless of the genotype.

The influence of the learning load on the interaction between learning and evolution is studied. It is shown that the learning load leads to the effective genetic assimilation and to a considerable acceleration of evolution.

These results were obtained for the main model as well as for the additional model that is based on the approach by Hinton and Nowlan (1987).

It should be underlined that our analysis essentially uses the quasispecies model (Eigen, 1971; Eigen and Schuster, 1979). Basing on this model, it is sufficient to consider only single significant variable, the distance to the optimum ρ. This ensures the clear understanding of mechanisms of interaction between learning and evolution.

**Acknowledgments**

This work is partially supported by the Russian Foundation for Basic Research, Grant No 13-01-00399.

## Appendix: Results of Estimation of Efficiency of Evolutionary Algorithms

The estimations (Red'ko and Tsoy, 2005, 2006) were made for the model of quasispecies (Eigen, 1971; Eigen and Schuster, 1979). This model describes the evolution of the population of organisms $\mathbf{S}_k$; each organism $\mathbf{S}_k$ is determined by the chain of symbols $S_{ki}$, symbols take two values: $S_{ki} = 0$ or $S_{ki} = 1$; $i = 1,2,\ldots,N$; $k = 1,2,\ldots,n$; $N$ is the length of chains; $n$ is the number of organisms in the population. The fitness of an organism $\mathbf{S}$ decreases exponentially with the Hamming distance $\rho(\mathbf{S},\mathbf{S_M})$ between $\mathbf{S}$ and the certain optimal chain $\mathbf{S_M}$ ($S_{Mi} = 0$ or $1$; $i = 1,2,\ldots,N$):

$$f(\mathbf{S}) = \exp[-\beta\rho(\mathbf{S},\mathbf{S_M})] , \qquad (5)$$

where $\beta$ is the parameter of selection intensity.

The evolutionary process consists of a number of generations; each generation consists of a) the selection of the organisms into the next generation that is performed by means of the method of fitness proportionate selection and b) the mutations that are random replacements of symbols $S_{ki}$. The probability of changing of any symbol in one generation at mutations is equal to $p_m$. The probability of the selection of a particular organism $\mathbf{S}$ into the new generation is proportional to its fitness $f(\mathbf{S})$. It is assumed that $N$, $n \gg 1$ and $2^N \gg n$ ($N$, $n$ = const). The initial population consists of random organisms, so the characteristic distance $\rho$ between the organisms $\mathbf{S}$ of this population and the optimal chain $\mathbf{S_M}$ is approximately equal to $N/2$.

New organisms having small values $\rho$ appear in the population owing to mutations and are fixed in the population by means of selection. The characteristic number of generations $G_{-1}$, which is needed to reduce the mean value $\rho$ in the population by 1, can be estimated as follows: $G_{-1} \sim G_m + G_s$. Here $G_m \sim (Np_m)^{-1}$ is the characteristic number of generations that is needed for mutations of organisms of the population, $G_s \sim \beta^{-1}$ is the characteristic number of generations that is needed for replacement of organisms, having $\rho = \langle\rho\rangle$, by more preferable organisms, having $\rho = \langle\rho\rangle-1$.

The total number of generations $G_T$ of the evolutionary process, which is needed for finding the optimal chain $\mathbf{S_M}$, is of the order of $G_T \sim G_{-1} N$, therefore, we have:

$$G_T \sim (p_m)^{-1} + N\beta^{-1}. \qquad (6)$$

Let us choose the parameters of the model for the given value $N$ in such a manner to minimize the total number of organisms participating in the evolutionary search for the optimal chain $\mathbf{S_M}$. We use the following assumptions.

1) The intensity of selection is enough large: $\beta \geq p_m N$; in this case we can neglect the second term in the expression (6), i.e., the speed of evolution is determined by the intensity of mutations.

2) The intensity of mutations must not be too large, in order to remove the possibility of mutational losses of already found successful organisms, and the intensity of mutations must not be too small in order to ensure rather quick evolutionary search for the optimal chain $\mathbf{S_M}$.

We believe that $p_m = N^{-1}$. Consequently, from (6) we estimate the total number of generations of the evolutionary search: $G_T \sim N$.

3) We assume the minimal allowable population size $n$, at which there are no significant losses of successful organisms as a result of the neutral selection. The characteristic number of generations of the neutral selection $G_n$ is of the order of the population size $n$ (Kimura, 1983; Red'ko and Tsoy, 2005): $G_n \sim n$. $G_n$ should be no less than $G_T$. Thus, the minimal allowable population size can be estimated as $n \sim G_T$.

Using these assumptions, we have $n \sim G_T \sim N$. Finally, we obtain estimations of the total number of generations of the evolutionary process $G_T$ and the total number of organisms involved in the evolutionary search $n_{total}$ ($n_{total} = n \, G_T$):

$$G_T \sim N, \quad n_{total} \sim N^2. \tag{7}$$

Computer simulations (Red'ko and Tsoy, 2005, 2006) confirmed the estimations (7). Thus, the parameters of the effective evolutionary search are: $n = N$, $p_m = N^{-1}$, $\beta = 1$.